%% file: root.tex
\documentclass[letterpaper, 10 pt, conference]{ieeeconf}  % Comment this line out if you need a4paper

\IEEEoverridecommandlockouts                              % This command is only needed if 
\usepackage{epsfig} % for postscript graphics files
\usepackage{amsmath} % assumes amsmath package installed
\usepackage{amssymb}  % assumes amsmath package installed

\usepackage{newtxtext}
\usepackage{newtxmath} 

\usepackage[bookmarks=true]{hyperref}
\usepackage{booktabs}
\usepackage{multirow}
\usepackage{graphicx}
\makeatletter
\def\endfigure{\end@float}
\def\endtable{\end@float}
\makeatother
\usepackage[caption=false,font=footnotesize]{subfig}
\makeatletter
\let\prism@orig@makecaption\@makecaption
\long\def\@makecaption#1#2{%
  \ifx\@captype\@IEEEtablestring
    \begin{center}{\footnotesize #1}. {\footnotesize #2}\end{center}%
    \@IEEEtablecaptionsepspace
  \else
    \prism@orig@makecaption{#1}{#2}%
  \fi
}
\makeatother
\usepackage{xcolor}
\usepackage{siunitx}
\usepackage{stfloats}
\usepackage{makecell}
\usepackage{float}
\usepackage{placeins}
\usepackage{xspace}
\usepackage{flushend}

\newcommand{\bpp}{\textsc{BEV-Patch-PF}\xspace}

\title{\LARGE \bf
\bpp: Particle Filtering with BEV-Aerial Feature Matching for Off-Road Geo-Localization
}

\author{
    Dongmyeong Lee$^{1}$, Jesse Quattrociocchi$^{2}$, Christian Ellis$^{2}$, Rwik Rana$^{1}$,\\
    Amanda Adkins$^{1}$, Adam Uccello$^{2}$, Garrett Warnell$^{2}$, and Joydeep Biswas$^{1}$\\[0.4em]
    \begin{tabular}{c@{\hspace{1.5em}}c}
        $^{1}$University of Texas at Austin
        &
        $^{2}$DEVCOM Army Research Laboratory
    \end{tabular}\\[0.25em]
    \url{https://amrl.cs.utexas.edu/bev-patch-pf}
}

\begin{document}

\maketitle
\thispagestyle{empty}
\pagestyle{empty}

\begin{abstract}
Localizing ground robots against aerial imagery provides a critical capability for autonomous navigation, especially in environments where GPS is unreliable or unavailable.
This task is challenging due to large viewpoint differences and substantial environmental variability.
Most prior methods localize each frame independently, using either global-descriptor retrieval or spatial feature alignment, which leaves them vulnerable to ambiguity and multi-modal pose hypotheses.
While sequential reasoning can mitigate this uncertainty, adapting existing per-frame pipelines for sequential use introduces unfavorable trade-offs among accuracy, memory, and computation that limit their practical deployment.
We propose \bpp, a vision-only, GPS-free sequential geo-localization system that integrates particle filtering with learned bird’s-eye-view (BEV) and aerial feature maps.
For each 3-DoF particle pose hypothesis, we crop the corresponding patch from an aerial feature map computed from a local aerial image centered on the predicted pose.
The resulting BEV–aerial feature match defines a per-particle log-likelihood for particle-filter updates.
In addition, we learn a frame-level uncertainty estimate that adaptively flattens the observation likelihood for unreliable observations, preventing overconfident particle collapse in ambiguous regions.
On two real-world off-road datasets, our method achieves \textit{9.7$\times$} lower absolute trajectory error (ATE) on seen routes and \textit{6.6$\times$} lower ATE on unseen routes than a retrieval-based baseline, while remaining robust under partial canopy cover and shadowing.
The system runs in real time at \textit{10 Hz} on an NVIDIA Tesla T4, enabling practical robot deployment.
\end{abstract}

\input{content/introduction}

\input{content/related_works}

\input{content/methodology}
\input{content/experiments}
\input{content/conclusion}

\FloatBarrier

\section*{Acknowledgments}
This work is partially supported by the ARL SARA (W911NF-24-2-0025 and W911NF-23-2-0211). Any opinions, findings, and conclusions expressed in this material are those of the authors and do not necessarily reflect the views of the sponsors.

\bibliographystyle{IEEEtran}
\bibliography{IEEEabrv,references}

\end{document}

%% file: content/introduction.tex
% Motivate the problem: 
% - Who cares about it and why? What are the challenges?
% - Cite the state of the art, and other papers if they have also revealed limitations of the SOTA.
% - Illustrate the limitations of the state of the art that you will be addressing.
% - Illustrate how your proposed solution overcomes the limitations, highlighting the key insights but without having to go into math.
% - Summarize your empirical findings.

\section{Introduction}

High-quality global localization in a geo-referenced frame allows robots to leverage aerial imagery, which can be used to provide improved long-range off-road planning and navigation around hazards such as cliffs and rivers.
Although visual and LiDAR-inertial odometry provide local pose estimates, they accumulate drift without global fixes, leading to errors that compromise downstream planning.

Cross-view geo-localization addresses the lack of global position fixes by estimating a robot’s 3-DoF pose in a UTM frame by matching ground-level images with geo-referenced aerial imagery. 
However, this task is inherently difficult due to potentially large viewpoint differences between the onboard and aerial sensors.
This problem is especially challenging in unstructured off-road environments, where the absence of man-made landmarks and the presence of terrain irregularities and tree canopy exacerbate the visual mismatch and remove many of the cues that conventional methods rely on~\cite{castaldo2015semantic, tian2017cross}.
Recent deep learning approaches typically tackle this problem frame-by-frame, falling into two main categories: \textit{retrieval-based} methods~\cite{zhu2021vigor, xia2021cross, xia2022visual, zhu2022transgeo, klammer2024bevloc} that learn global descriptors for ground and aerial images, and \textit{spatial feature-alignment} methods~\cite{sarlin2023orienternet, fervers2023uncertainty, shi2022beyond, song2023learning, shi2023boosting} that infer poses by aligning features in a shared representation.
Per-frame localization, however, considers only a single observation at a time, making it vulnerable to ambiguity and multi-modal solutions.
In off-road settings, this can lead to catastrophic pose jumps caused by visually similar map regions or sensor occlusions.
Sequential localization mitigates these issues by enforcing temporal consistency.

\input{figures_tex/thumbnail}

While sequential inference can reduce pose ambiguity, it requires observation models that yield smooth, discriminative likelihoods over continuous pose hypotheses. 
Most prior cross-view methods~\cite{xia2022visual, zhu2022transgeo, sarlin2023orienternet, fervers2023uncertainty} do not provide continuous likelihoods.
Retrieval-based approaches assign similarity scores over a discretized set of aerial patches, making them insensitive to fine-grained pose changes and unsuitable for continuous probabilistic filtering.
In contrast, spatial feature-alignment methods offer improved granularity but they either: 
(i) require dense correlation over discretized pose grids\textemdash incurring high computational cost or
(ii) optimize a single best pose, which is difficult to use as a likelihood over hypotheses.

To address these limitations, we introduce \bpp, a sequential localization system that integrates a particle filter with an observation model evaluating likelihoods over continuous pose. 
From onboard RGB/depth images, we construct a bird's-eye view (BEV) feature map; for each particle pose, we extract the corresponding aerial feature patch and compare it to the BEV features to obtain a per-particle log-likelihood.
Because aerial patches can be sampled at arbitrary continuous poses, the likelihood is computed directly at each particle hypothesis, making the approach a natural fit for particle filtering. 
The model targets unstructured off-road terrain and does not rely on explicit semantic landmarks.

We evaluate our approach on real-world off-road datasets, including TartanDrive~\cite{sivaprakasam2024tartandrive} and a new dataset called \textsc{UT-SARA-GQ}, which we introduce to specifically test performance under tree canopy.
We compare against a retrieval-based pose-graph-optimization method~\cite{klammer2024bevloc} and visual / LiDAR / wheel odometry systems~\cite{PyCuVSLAM, zhao2021super}.
Across seen and unseen routes from the TartanDrive 2.0~\cite{sivaprakasam2024tartandrive} dataset, our method consistently achieves lower trajectory error and greater robustness.
These results demonstrate the benefits of modeling continuous-pose likelihoods and confirm generalization to previously unobserved routes.

In summary, our contributions are as follows:
\begin{enumerate}
    \item A novel observation model for particle filtering that computes continuous-pose likelihoods by matching learned BEV features from ground RGB-D images to features from an aerial orthophoto.
    \item Strong performance on off-road localization, with extensive experiments showing significant accuracy gains over existing methods and robust generalization to routes not seen during training.
    \item A new public \textsc{UT-SARA-GQ} dataset and benchmark for evaluating cross-view localization under challenging canopy and shadow occlusions, along with experiments validating our method's robustness.
    \item A real-time and deployment-ready system, including an open-source C++ ROS 2 wrapper with a TensorRT-optimized inference engine for practical field robotics.
\end{enumerate}

%% file: figures_tex/thumbnail.tex
\begin{figure}[t]
  \centering

  % Row 1
  \subfloat[\label{fig:1a}]{%
    \includegraphics[width=0.32\linewidth]{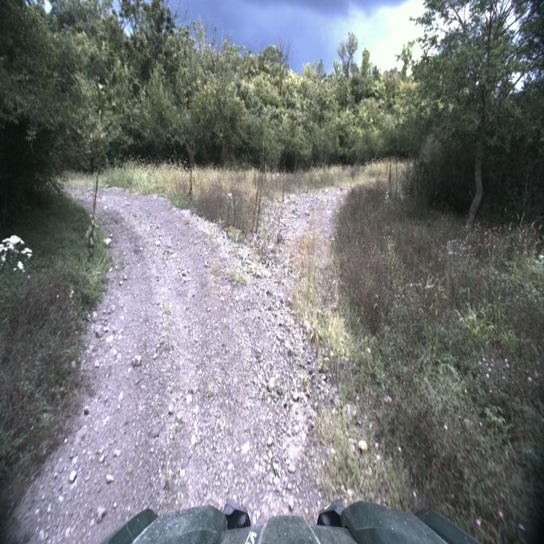}}%
  \hfill
  \subfloat[\label{fig:1b}]{%
    \includegraphics[width=0.32\linewidth]{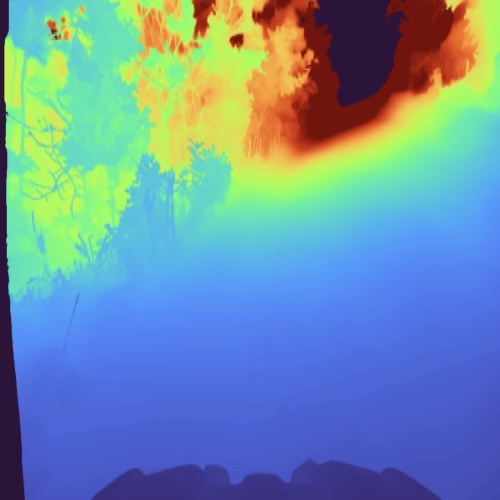}}%
  \hfill
  \subfloat[\label{fig:1c}]{%
    \includegraphics[width=0.32\linewidth]{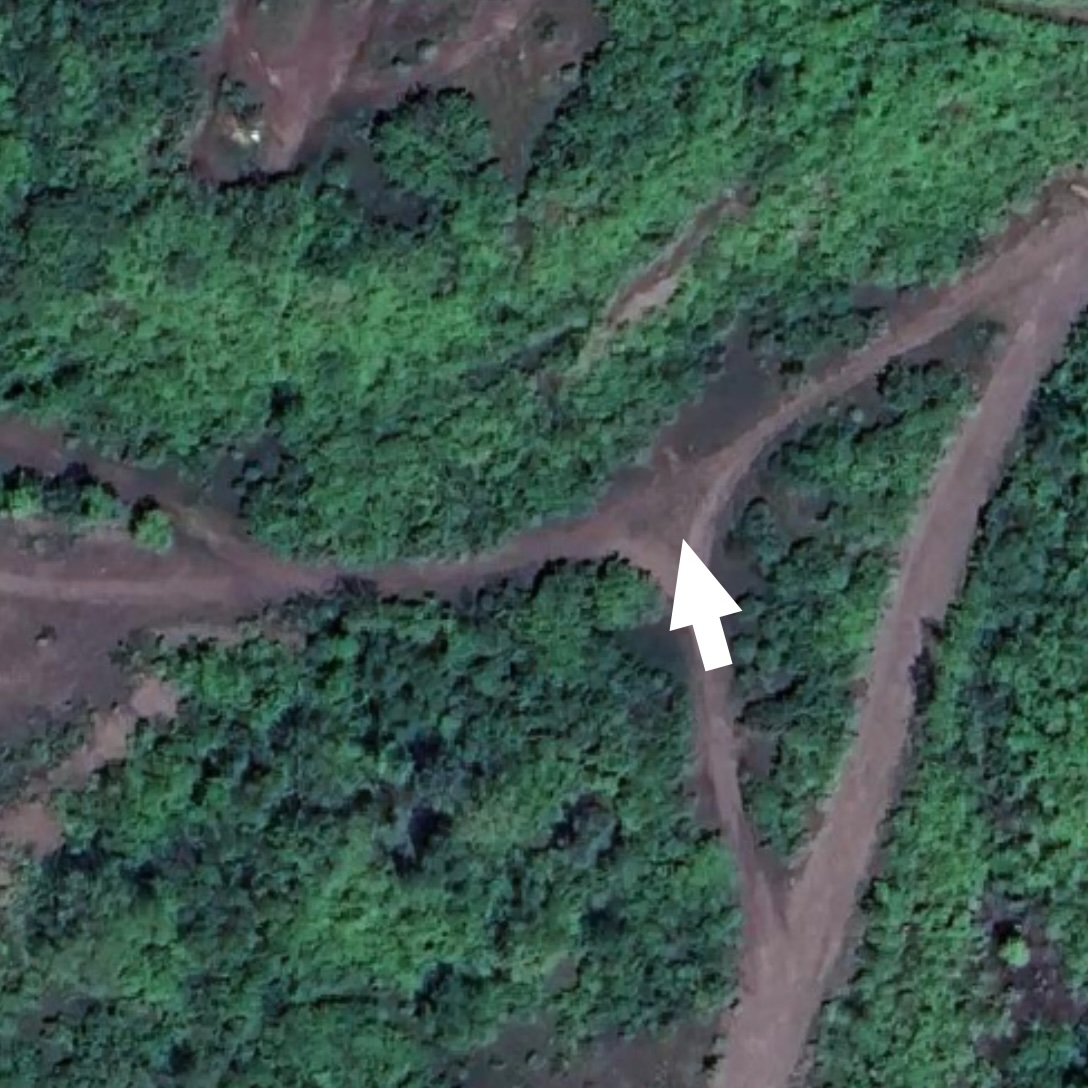}}%
  \\

  % Row 2
  \subfloat[\label{fig:2a}]{%
    \includegraphics[width=0.32\linewidth]{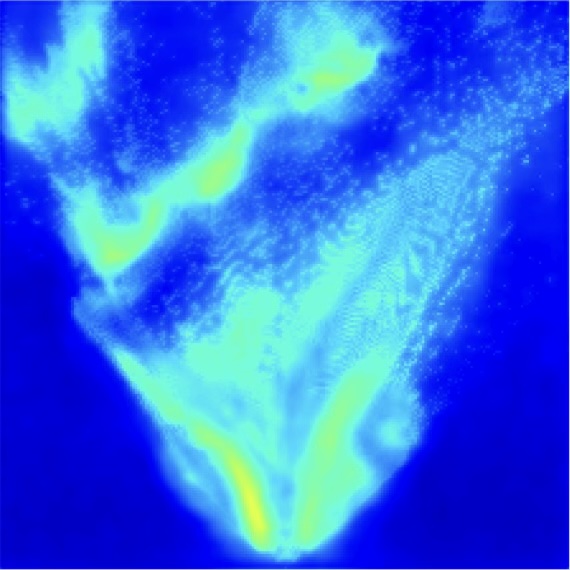}}%
  \hfill
  \subfloat[\label{fig:2b}]{%
    \includegraphics[width=0.32\linewidth]{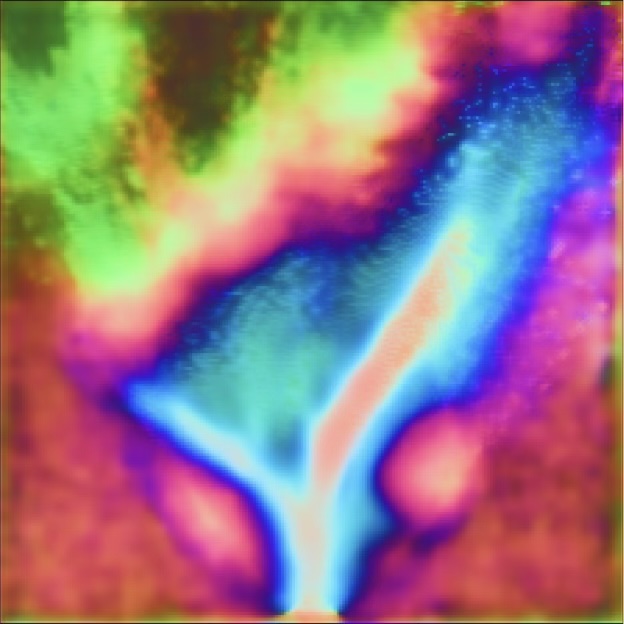}}%
  \hfill
  \subfloat[\label{fig:2c}]{%
    \includegraphics[width=0.32\linewidth]{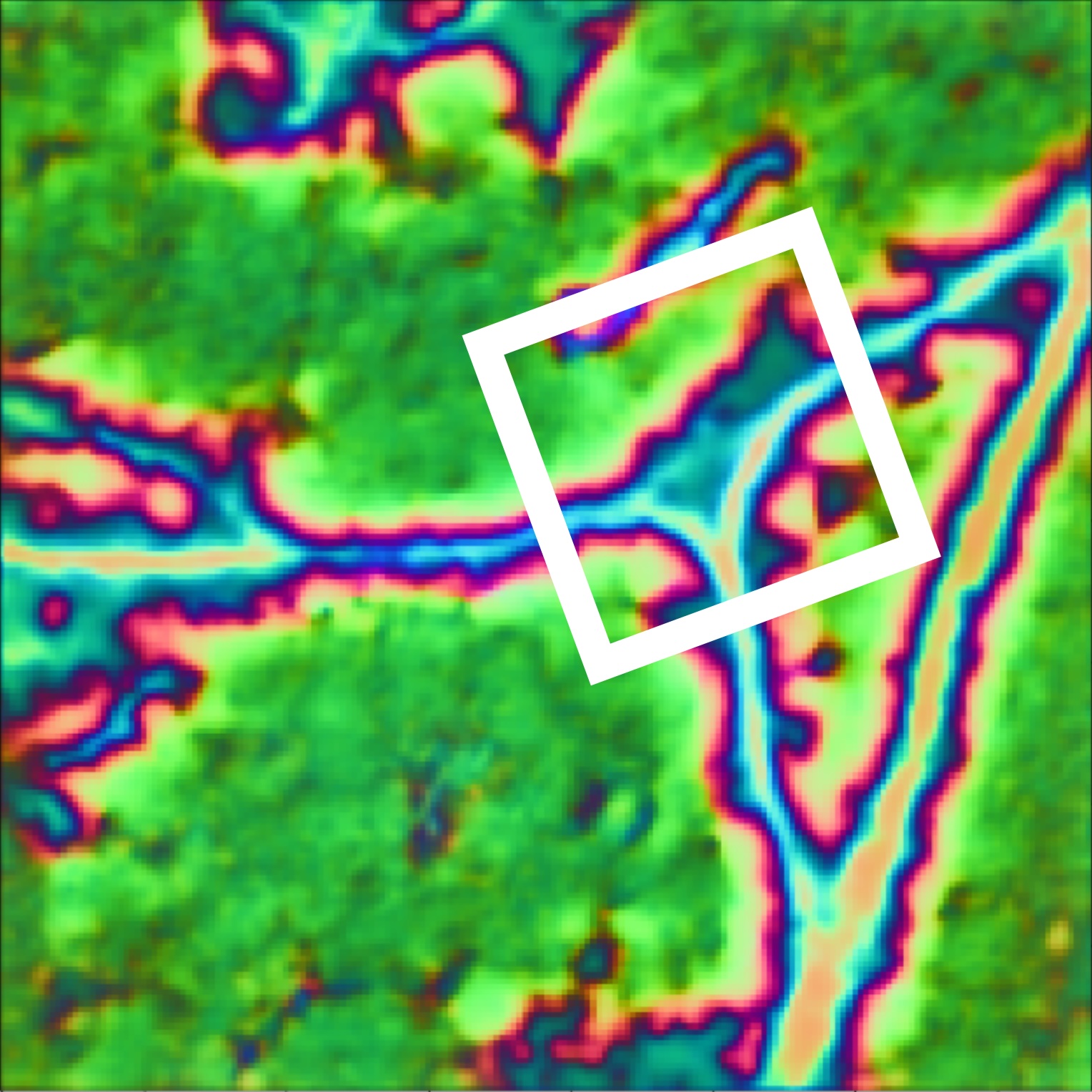}}%

  \caption{Visualization of \bpp inputs and outputs. \textbf{Top (Inputs):}
  (a) onboard RGB image $\mathcal{I}$, %
  (b) depth image $\mathcal{D}$, and %
  (c) a local aerial orthophoto $\mathcal{M}[\hat{\mathbf{x}}_{t|t-1}]$, where the white arrow indicates the ground-truth pose. %
  \textbf{Bottom (Outputs):} (d) the BEV distinctiveness map $\mathbf{C}$, %
  (e) the corresponding feature map $\mathbf{G}$, and %
  (f) the aerial feature map $\mathbf{F}$. %
  The white box in (f) illustrates one representative sampled patch; during inference, \bpp evaluates a patch for each particle hypothesis.}
  \label{fig:thumbnail}
\end{figure}

%% file: content/related_works.tex
\section{Related Works}
\label{sec:related-works}

Visual geo-localization aims to estimate a robot's 3-DoF pose within a geo-referenced map using ground-level imagery.
A classic formulation is that of visual place recognition (VPR), where a query image is matched against a pre-collected database of geo-tagged images to find the closest corresponding location~\cite{keetha2023anyloc, lu2024cricavpr}.
While effective in densely mapped urban areas, VPR is often impractical for off-road missions where comprehensive prior data collection is not feasible.

\noindent \textbf{Cross-view geo-localization with single frames:}
To overcome the need for a ground-level database, cross-view geo-localization methods match ground images directly to overhead imagery, such as satellite photos or planimetric maps. Early deep-learning approaches focused on learning cross-view descriptors \cite{zhu2021vigor, xia2022visual, zhu2022transgeo, klammer2024bevloc}. These methods typically use contrastive learning to align the embedding of a ground image with that of its corresponding aerial patch. However, their accuracy is often limited by the discretization of the aerial map and a lack of explicit orientation modeling. While later work began to infer heading by encoding multiple rotations per grid cell \cite{xia2023convolutional, lentsch2023slicematch}, these estimates remain coarse.

To achieve finer pose granularity, spatial-feature-alignment methods were introduced. These techniques, which include dense cross-correlation in BEV space \cite{sarlin2023orienternet, fervers2023uncertainty} and continuous-pose optimization \cite{shi2022beyond, song2023learning, shi2023boosting}, directly align learned features from both views. Dense correlation methods attain high precision but require sweeping $K$ rotations over an $H{\times}W$ grid, resulting in a computational complexity of $\mathcal{O}(K H^{2}W^{2})$. Conversely, continuous optimization avoids this exhaustive search but is susceptible to converging in local minima.

\noindent \textbf{Sequential estimation for temporal consistency:}
Per-frame methods are fundamentally challenged by multi-modality and a lack of temporal consistency.
Their reliance on single-frame observations provides no mechanism to distinguish between visually similar locations or to ensure the final trajectory is smooth and logical over time.
To address this, sequential methods enforce chronological consistency.
For instance, OrienterNet~\cite{sarlin2023orienternet} warps dense probability maps over time but requires ground-truth odometry.
BEVLoc~\cite{klammer2024bevloc} embeds per-frame localizations into a pose graph but requires approximate absolute position fixes (e.g., GPS) to filter outliers.
Similarly, a recent end-to-end particle smoother~\cite{younis2024learning} shows strong performance but is confined to urban scenes with planimetric maps.

A common approach for sequential inference is to combine retrieval with a particle filter (PF) \cite{xia2021cross, hu2020image, zhou2021efficient}.
In these systems, each particle represents a pose hypothesis and queries the descriptor of the nearest map cell for comparison with the ground-view descriptor.
However, this technique inherits the limitations of retrieval-based methods, namely its dependence on grid discretization and coarse yaw bins, which blurs the likelihood distribution over a continuous pose space.

\noindent \textbf{Off-road cross-view localization:}
Despite these advances, the problem of off-road cross-view geo-localization remains under-explored.
The vast majority of existing methods and datasets focus on structured urban scenes~\cite{zhu2021vigor, xia2021cross, sarlin2023orienternet, agarwal2020ford}.
These environments provide strong structural cues, such as buildings and roads, and often include semantic map annotations that are absent in unstructured terrain.
The visual challenges of off-road environments, including texture-poor ground, dense vegetation, and irregular terrain, render the assumptions underlying urban-centric methods untenable.
To our knowledge, only BEVLoc~\cite{klammer2024bevloc} and BEVRender~\cite{jin2024bevrender} have conducted experiments in off-road settings, which suggests that robust localization for off-road environments remains underexplored.

%% file: content/methodology.tex
\section{\texorpdfstring{Particle Filtering with\\ BEV--Aerial Feature Matching}
{Particle Filtering with BEV-Aerial Feature Matching}}

\begin{figure*}[htb]
    \centering
    \includegraphics[width=0.7\linewidth]{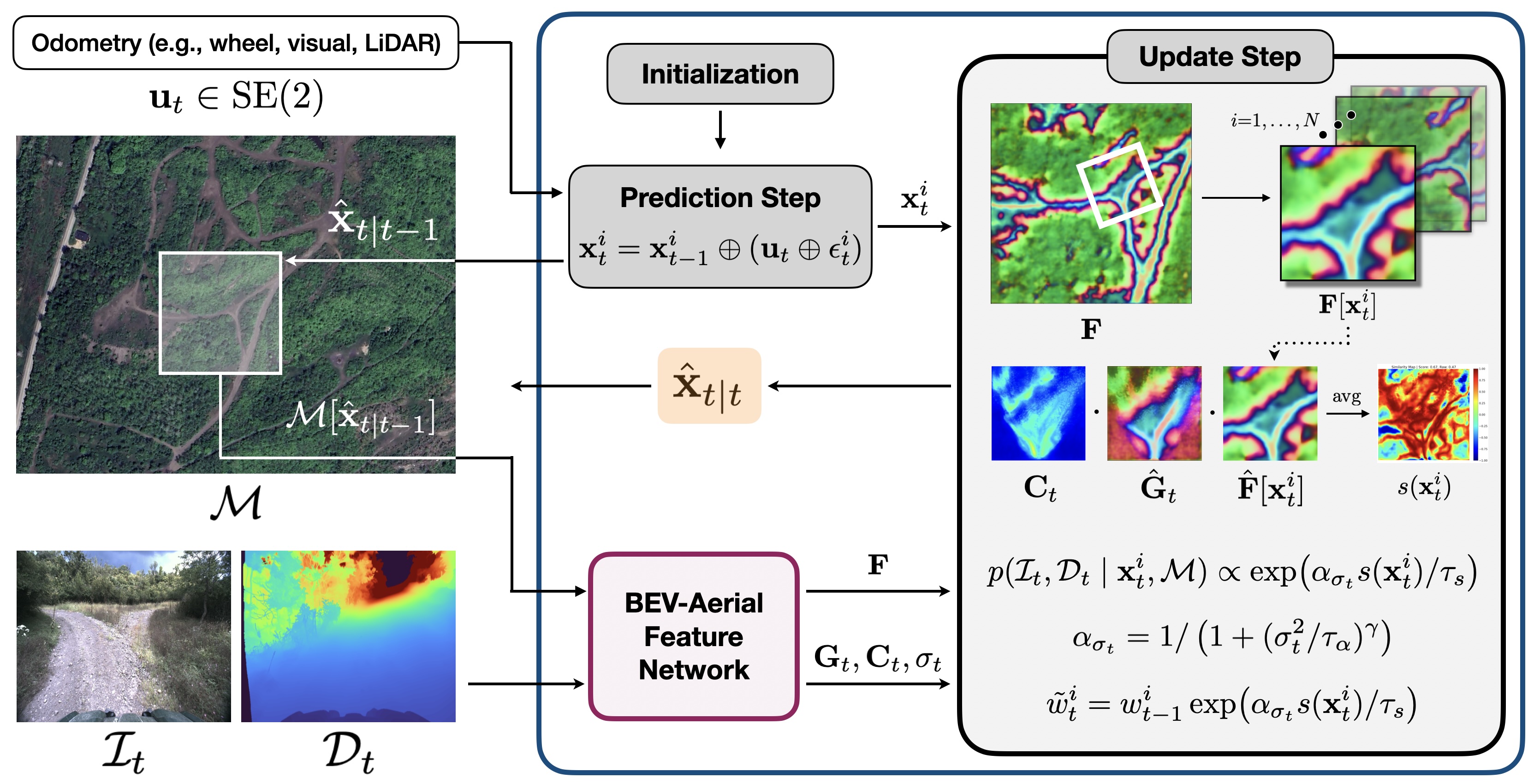}
    \caption{Overall pipeline of the \bpp. The heatmap in the update step visualizes the weighted similarity used to compute the matching score $s(\mathbf{x}^i_t)$.}
    \label{fig:system-pipeline}
\end{figure*}

We propose \bpp, a sequential localization framework that combines particle filtering with a learned observation model based on bird's-eye-view (BEV) and aerial feature matching.
At each time step, particles are reweighted using a similarity score between the onboard BEV representation and the corresponding aerial feature patch sampled at each particle pose.
This recursive filtering loop enables robust localization in challenging environments.

%%%%%%%%%%%%%%%%%%%%%%%%%%%%%%%%%%%%%%%%%%%%%%%%%%%%%%%
% Problem Formulation
%%%%%%%%%%%%%%%%%%%%%%%%%%%%%%%%%%%%%%%%%%%%%%%%%%%%%%%
\subsection{Problem Formulation}
Our objective is to recursively estimate the ground robot's 3-DoF pose $\mathbf{x}_t = (x_t, y_t, \theta_t) \in \mathrm{SE}(2)$, where $(x_t, y_t)$ are east- and north-directed UTM coordinates (in meters) and $\theta_t$ is the heading, measured counter-clockwise from the east axis of a north-up satellite map.

At each timestamp $t$, the system receives an onboard observation $z_t = \{\mathcal{I}_t, \mathcal{D}_t\}$, consisting of an RGB image and its corresponding depth map, along with a relative motion estimate $\mathbf{u}_t \in \mathrm{SE}(2)$ from an odometry source.

The main problem is to estimate the belief distribution $p(\mathbf{x}_t \mid z_{1:t}, \mathbf{u}_{1:t}, \mathcal{M})$ using a particle filter. 
The distribution is represented as a set of $N$ weighted samples, $\{(\mathbf{x}^i_t, w^i_t)\}^{N}_{i=1}$, with each particle $\mathbf{x}^i_t$ denoting a discrete state hypothesis and $w^i_t$ its associated weight.
The particle set is updated via Bayesian filtering~\cite{thrun2002probabilistic}.
Each particle is reweighted according to the likelihood of the current observation $z_t$:
\begin{equation}
    w^i_t \propto w^i_{t-1} \, p(z_t \mid \mathbf{x}^i_t, \mathcal{M})
\end{equation}
This reweighting process, paired with a motion prediction step, allows the filter to recursively refine its estimate.
Our specific implementation of these steps is detailed next.

%%%%%%%%%%%%%%%%%%%%%%%%%%%%%%%%%%%%%%%%%%%%%%%%%%%%%%%
% Particle Filter Localization
%%%%%%%%%%%%%%%%%%%%%%%%%%%%%%%%%%%%%%%%%%%%%%%%%%%%%%%
\subsection{Particle Filter Localization}
\label{subsec:particle-filter}
The overall \bpp pipeline is illustrated in Fig.~\ref{fig:system-pipeline}.

\noindent \textbf{Initialization:}
Initially, a set of $N$ particles is distributed with Gaussian noise around a coarse initial pose, which can be provided through manual selection.
The filter then enters the recursive prediction and update cycle.

\noindent \textbf{Prediction step:}
Each particle pose $\mathbf{x}^i_t$ at time $t$ is obtained from its predecessor $\mathbf{x}^i_{t-1}$ by propagating the motion estimate $\mathbf{u}_t$ and adding Gaussian noise to account for odometry error:
\begin{equation}
    \mathbf{x}^i_t = \mathbf{x}^i_{t-1} \oplus (\mathbf{u}_t \oplus \epsilon^i_t),
    \quad
    \epsilon^i_t = \operatorname{Exp}(\boldsymbol{\delta}^i_t).
\end{equation}
Here, the operator $\oplus$ denotes composition on the $\mathrm{SE}(2)$ group, and the process noise $\epsilon^i_t$ is generated by sampling a vector $ \boldsymbol{\delta}^i_t \in \mathbb{R}^3$ from a zero-mean Gaussian distribution whose covariance is proportional to the odometry $\mathbf{u}_t$, and then mapping it from the Lie algebra $\mathfrak{se}(2)$ to the group $\mathrm{SE}(2)$ via the exponential map $\operatorname{Exp}(\cdot)$.

\noindent \textbf{Update step:}
At each time $t$, we update the particle weights $w^i_t$ based on the measurement likelihood $p(z_t \mid \mathbf{x}^i_t, \mathcal{M})$.

First, the BEV--aerial feature network (described in Sec.~\ref{subsec:bev-aer-feat-net}) computes a BEV feature map $\mathbf{G} \in \mathbb{R}^{H_b \times W_b \times D}$, a distinctiveness map $\mathbf{C} \in {[0,1]}^{H_b \times W_b}$, and a frame-level uncertainty $\sigma_t$ from the onboard RGB-D image.

Second, for computational efficiency, we avoid processing a separate aerial crop for each of the $N$ particles.
Instead, we extract a single larger aerial crop $\mathcal{M}[\hat{\mathbf{x}}_{t|t-1}]$ from the global aerial map $\mathcal{M}$, centered at a representative predicted pose $\hat{\mathbf{x}}_{t|t-1}$ computed from the predicted particle set.
This local aerial image is then processed by the network to produce an aerial feature map $\mathbf{F} \in \mathbb{R}^{H_a \times W_a \times D}$.

Third, for each particle hypothesis $\mathbf{x}^i_t$, we sample its corresponding patch $\mathbf{F}[\mathbf{x}^i_t]$ from the aerial feature map $\mathbf{F}$ via bilinear sampling.
An affine sampling grid is constructed for each particle, rotated by its heading and anchored at its position, which corresponds to the bottom-center of the patch.
This grid is then used to sample a $H_b \times W_b$ patch that is spatially aligned with the BEV feature map $\mathbf{G}$.
This approach assumes that the particle distribution is compact enough to remain mostly within the local aerial image $\mathcal{M}[\hat{\mathbf{x}}_{t|t-1}]$.
To handle outlier particles that may fall outside this boundary, the sampler uses zero-padding for any out-of-bounds coordinates.

Finally, we compute a matching score $s(\mathbf{x}^i_t) \in [-1, 1]$ for each particle by comparing the BEV feature $\mathbf{G}$ with the sampled aerial feature patch $\mathbf{F}[\mathbf{x}^i_t]$.
The distinctiveness map $\mathbf{C}$ assigns higher weights to spatial locations that are more informative for discriminating the correct pose from incorrect pose hypotheses.
We compute the score as the mean distinctiveness-weighted cosine similarity over the BEV grid:
\begin{equation}
    s(\mathbf{x}^i_t)
    =
    \frac{1}{H_bW_b}
    \sum_{v=1}^{H_b}\sum_{u=1}^{W_b}
    \mathbf{C}_{uv}
    \left( \hat{\mathbf{G}}_{uv}^{\top} \, \hat{\mathbf{F}}[\mathbf{x}^i_t]_{uv} \right).
\end{equation}
Here, $\hat{\mathbf{G}}$ and $\hat{\mathbf{F}}$ denote the corresponding $\ell_2$-normalized BEV and aerial feature maps, respectively.

This matching score $s(\mathbf{x}^i_t)$ is then converted to an observation likelihood, which represents the probability of the current observation given the particle's pose:
\begin{equation}
    p(z_t \mid \mathbf{x}^i_t, \mathcal{M})
    \propto
    \exp\!\left( \alpha_{\sigma_t} \, s(\mathbf{x}^i_t) / \tau_s \right),
\end{equation}
\begin{equation}
    \alpha_{\sigma_t} = \frac{1}{1+(\sigma^2_t/\tau_\alpha)^\gamma}.
\end{equation}
Here, $\tau_s$ is a temperature hyperparameter that controls the sharpness of the likelihood distribution, whereas $\tau_\alpha$ and $\gamma$ modulate how the frame-level uncertainty $\sigma_t$ attenuates the distribution to prevent overconfident updates from uncertain observations.
Particle weights are then updated as
\begin{equation}
    \tilde{w}^i_t = w^i_{t-1} \exp\!\left( \alpha_{\sigma_t} s(\mathbf{x}^i_t) / \tau_s \right)
\end{equation}
and subsequently normalized to obtain $w^i_t$.

\noindent \textbf{Resampling step:}
Low-variance resampling is triggered only when the effective sample size falls below a preset threshold, preserving particles most consistent with the true pose while discarding less plausible hypotheses.

%%%%%%%%%%%%%%%%%%%%%%%%%%%%%%%%%%%%%%%%%%%%%%%%%%%%%%%
% BEV-Aerial Feature Network
%%%%%%%%%%%%%%%%%%%%%%%%%%%%%%%%%%%%%%%%%%%%%%%%%%%%%%%
\subsection{BEV--Aerial Feature Network}
\label{subsec:bev-aer-feat-net}

\begin{figure}[htb]
    \centering
    \includegraphics[width=0.9\linewidth]{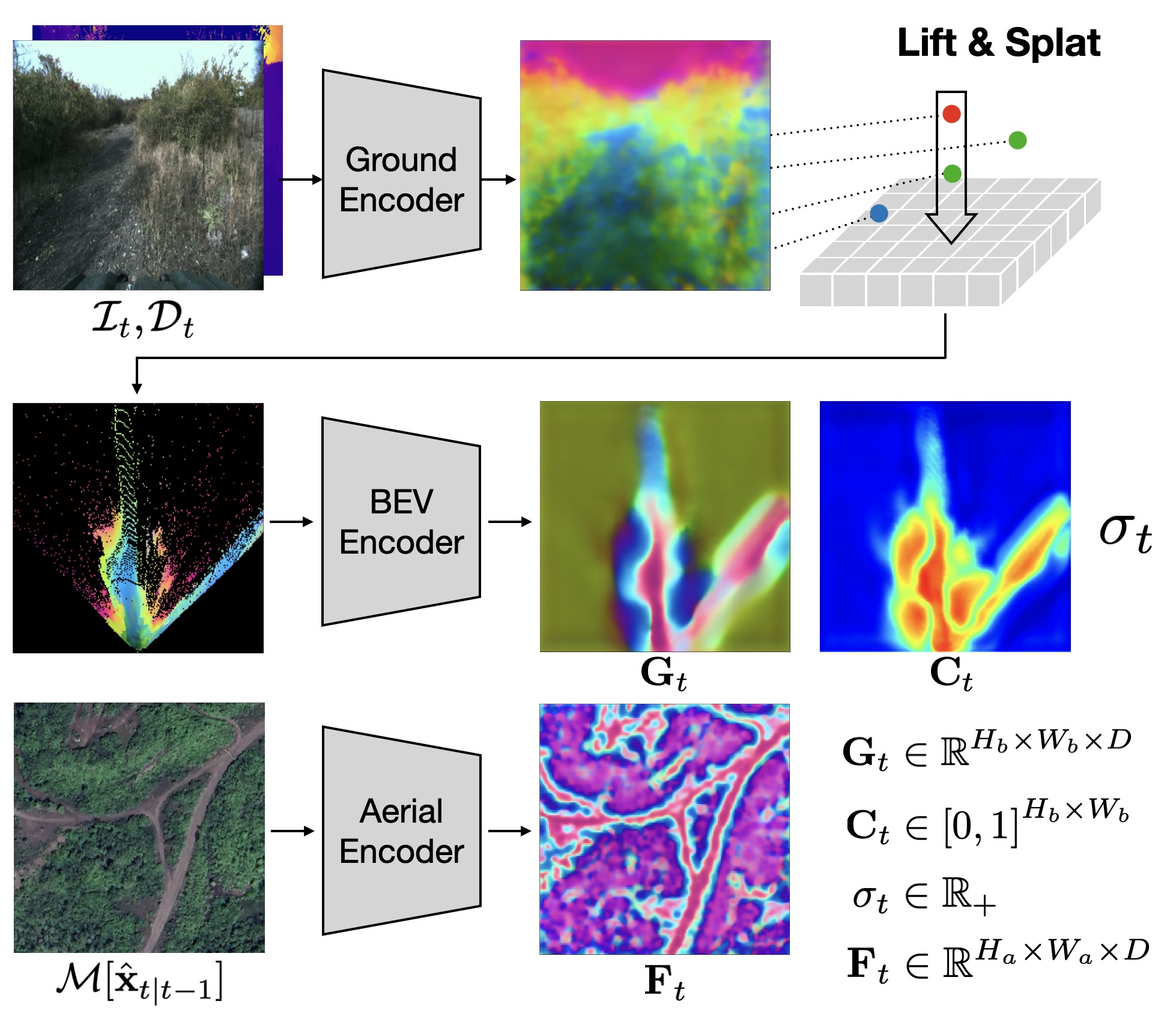}
    \caption{BEV-Aerial feature network architecture.}
    \label{fig:bev-network}
\end{figure}

The core of our observation model is a feature network that produces the BEV features $\mathbf{G}$, the BEV distinctiveness map $\mathbf{C}$, frame-level uncertainty $\sigma_t$, and the local aerial feature map $\mathbf{F}$.
As illustrated in Fig.~\ref{fig:bev-network}, the network contains several modules.

\noindent \textbf{Ground encoder:}
The ground encoder takes the onboard RGB image $\mathcal{I}_t$ and extracts a feature map.
We employ a frozen, pre-trained DINOv3-ConvNeXt-Tiny~\cite{simeoni2025dinov3} visual foundation model for its general-purpose feature extraction capabilities. 
The output features are processed with UPerNet~\cite{xiao2018unified} to obtain a higher-resolution feature map, which is then fed to the BEV mapper.

\noindent \textbf{BEV mapper:}
The 2D features from the ground encoder are projected into a 2D BEV representation by the BEV Mapper.
Following the Lift-Splat~\cite{philion2020lift} methodology, we use the depth image $\mathcal{D}_t$ to back-project image features into 3D points in the robot's coordinate frame.
To maintain memory efficiency, we avoid creating a dense voxel grid and instead flatten the 3D points into a 2D BEV grid.
This grid represents a fixed-size area in front of the robot, defined in its local coordinate frame.
The grid's resolution is set to match that of the satellite map.
For each BEV grid cell, we compute a height-invariant weighted average of all 3D point features that fall within its vertical column.
The weight for each point feature is estimated by a two-layer MLP, allowing the network to prioritize more informative points during splatting.

\noindent \textbf{BEV encoder:}
The splatted BEV representation is then refined by a BEV Encoder.
This module consists of three sequential residual blocks followed by a UPerNet~\cite{xiao2018unified} head to aggregate spatial context.
The encoder outputs a tensor of shape $\mathbb{R}^{H_b \times W_b \times (D+2)}$.
The first $D$ channels form the BEV descriptor map $\mathbf{G}$.
The $(D{+}1)$-th channel produces pixel-wise distinctiveness logits, and applying a sigmoid yields the distinctiveness map $\mathbf{C}$.
The final channel produces an uncertainty evidence map, whose masked spatial average over valid BEV cells yields a scalar frame-level uncertainty $\sigma_t$.

\noindent \textbf{Aerial encoder:}
The aerial encoder processes the cropped satellite image $\mathcal{M}[\hat{\mathbf{x}}_{t|t-1}]$ using a DINOv3-ConvNeXt-Tiny backbone and a lightweight multi-scale decoder to produce the aerial feature map $\mathbf{F}$.
Unlike the ground encoder, which uses a UPerNet head, the aerial encoder uses a substantially lighter decoder that fuses intermediate backbone features with simple projection, upsampling, and shallow convolutional refinement.
This design better preserves local pixel-level consistency and empirically reduces blob-like degradation during training.

%%%%%%%%%%%%%%%%%%%%%%%%%%%%%%%%%%%%%%%%%%%%%%%%%%%%%%%
% Training Objective
%%%%%%%%%%%%%%%%%%%%%%%%%%%%%%%%%%%%%%%%%%%%%%%%%%%%%%%
\subsection{Training Objective}
\label{subsec:training-objective}

The training objective is designed to learn \textit{(i)} discriminative BEV--aerial matching features, \textit{(ii)} a pixel-wise distinctiveness map for spatial weighting.

\noindent \textbf{Uncertainty-weighted matching loss:}
To learn a discriminative feature representation, we use an InfoNCE loss over the candidate aerial patches.
For each training sample, the ground-truth (GT) pose $\mathbf{x}^{+}$ serves as the positive, while sampled poses $\mathcal{X}^{-}$ around the GT pose serve as negatives.
We jointly learn a frame-level uncertainty $\sigma_t$ and use it to reweight and regularize the matching loss:
\begin{equation}
    \mathcal{L}_\text{match}
    = 
    \frac{\mathcal{L}_\text{sim}}{\sigma^2_t} + \log \sigma^2_t,
\end{equation}
\begin{equation}
    \mathcal{L}_\text{sim}
    =
    -\log \frac{\exp(s(\mathbf{x}^{+}) / \tau)}{\exp(s(\mathbf{x}^{+}) / \tau) + \sum_{\mathbf{x} \in \mathcal{X}^{-}} \exp(s(\mathbf{x}) / \tau)}.
\end{equation}
This encourages the model to assign larger uncertainty to difficult observations, such as feature-poor open areas or heavy vegetation.
When computing this loss, we stop gradients through the distinctiveness map to prevent the network from trivially increasing distinctiveness weights.

\noindent \textbf{Distinctiveness loss:}
The distinctiveness map $\mathbf{C}$ is trained in a self-supervised manner to predict which spatial locations are informative for distinguishing the correct pose from incorrect pose hypotheses.
Let $\phi_{uv}(\mathbf{x}) = \hat{\mathbf{G}}_{uv}^{\top}\hat{\mathbf{F}}[\mathbf{x}]_{uv}$ denote the per-pixel cosine similarity for pose $\mathbf{x}$.
Among the sampled negative poses, we select the hardest negative, $\mathbf{x}^{-}_{\mathrm{hard}} = \arg\max_{\mathbf{x}\in\mathcal{X}^{-}} s(\mathbf{x})$.
The target distinctiveness map is then defined as follows:
\begin{equation}
\begin{aligned}
    \mathbf{C}^{*}_{uv}
    &=
    \operatorname{sigmoid}\bigl(c_p \, \phi_{uv}(\mathbf{x}^{+})\bigr)\cdot \\
    &\quad
    \operatorname{sigmoid}\Bigl(
        c_m \,
        \bigl[
            \phi_{uv}(\mathbf{x}^{+}) - \phi_{uv}(\mathbf{x}^{-}_{\mathrm{hard}})
        \bigr]
    \Bigr),
\end{aligned}
\end{equation}
where $c_p$ and $c_m$ control the influence of the positive-pose similarity and the similarity margin, respectively.
The target $\mathbf{C}^{*}$ is high at pixels that match the positive patch well while also separating it from the hardest negative patch.
A binary cross-entropy (BCE) loss then trains the predicted distinctiveness map $\mathbf{C}$ to match this target:
\begin{equation}
    \mathcal{L}_\text{distinct} = \operatorname{BCE} \bigl(\mathbf{C}, \mathbf{C}^{*}\bigr).
\end{equation}

The final training objective is
\begin{equation}
    \mathcal{L}_{\text{total}}
    =
    \mathcal{L}_{\text{match}}
    +
    \lambda_{\text{d}} \, \mathcal{L}_{\text{distinct}}
\end{equation}

%% file: content/experiments.tex
% Recipe for the experimental results section:
% First, list the questions you seek to answer empirically
% Optionally, have a subsection to describe the setup if it is involved, and is common to all or most of the experiments.
% Next, have subsections for each question, in the order of the list.
% In each sub-section:
% Explain the setup
% Explain the metrics
% Explain the comparison baselines
% Present the figure / table, and summarize the key takeaways.
% Note your conclusions.

\section{Experiments}
\label{sec:experiments}

We evaluate \bpp on offline datasets and real-time onboard deployments to answer four key questions:
\begin{enumerate}
    \renewcommand{\labelenumi}{Q\theenumi)}
    \item \textbf{Tracking accuracy:} How accurately does our method localize across different platforms and environments?
    \item \textbf{Generalization:} How robustly does the method perform on routes not seen during training?
    \item \textbf{Canopy/shadow robustness:} How reliably does the method localize under tree canopy cover and shadowing?
    \item \textbf{Real-time performance:} Does the system meet onboard compute and latency constraints for real-time operation?
\end{enumerate}

%%%%%%%%%%%%%%%%%%%%%%%%%%%%%%%%%%%%%%%%%%%%%%%%%%%%%%%
% Experimental Setup
%%%%%%%%%%%%%%%%%%%%%%%%%%%%%%%%%%%%%%%%%%%%%%%%%%%%%%%
\subsection{Experimental Setup}
\label{subsec:experimental-setup}

\noindent \textbf{Datasets:}
We evaluate on three challenging off-road datasets:
\begin{enumerate}
    \item TartanDrive~2.0~\cite{sivaprakasam2024tartandrive}: Collected with an ATV, this dataset includes 58 trajectories, which we split into 27 for training, 9 for validation, and 22 for testing. The test set is partitioned into 6 seen routes (overlapping training paths) and 16 unseen routes (novel paths).
    \item \textsc{UT-SARA-GQ}: A dataset we collected with a Clearpath Warthog in areas with tree-canopy cover and strong shadowing. Totaling 8.3~km and 60k frames, the dataset consists of 15 trajectories, split into 9 for training, 2 for validation, and 4 for testing.
    \item Urban park: To evaluate real-time performance, we collected an additional dataset in a local urban park. It contains 5 trajectories, which we split into 2 for training, 1 for validation, and 2 for testing.
\end{enumerate}

\noindent \textbf{Georeferenced imagery:}
For all experiments, we use north-up RGB satellite orthophotos (GeoTIFFs).
To improve model robustness, we train using a dynamically sampled image resolution ranging from \qtyrange{0.15}{0.45}{m/px}.
For evaluation, we use a fixed resolution of \qty{0.3}{m/px}.
All imagery was reprojected to the appropriate UTM zone using \textsc{QGIS}~\cite{QGIS_software}.%
\footnote{We reproject Google Satellite imagery to the target UTM zones: 17N (TartanDrive~2.0), 18N (UT-SARA-GQ dataset), and 14N (Urban Park).}

\noindent \textbf{Baselines:}
We compare our method against the following baselines.
For offline evaluation, all methods are initialized with the ground-truth starting pose of each trajectory to isolate drift accumulation. Note that \bpp uses no GPS/GNSS or other absolute position fixes during inference.

\begin{enumerate}
    \item BEVLoc~\cite{klammer2024bevloc}: %
    A recent cross-view localization method. We retrained the official code on our data splits and used stereo visual odometry as its motion prior. Following its original design, BEVLoc periodically applies absolute position fixes to regularize the pose graph; for TartanDrive, we provide the ground-truth positions available in the dataset as oracle position fixes, together with ground-truth orientation for heading assistance.
    \item PyCuVSLAM~\cite{korovko2025cuvslamcudaacceleratedvisual}: %
    A high-performance stereo visual odometry baseline.
    \item Super Odometry~\cite{zhao2021super}: %
    A LiDAR-Inertial odometry baseline using the pre-computed trajectories provided with TartanDrive~2.0.
\end{enumerate}

\noindent \textbf{Evaluation metrics:}
We report Absolute Trajectory Error (ATE) in meters, computed as the root-mean-square error (RMSE) between the estimated and ground-truth trajectories in the UTM coordinate frame.
No post-alignment is performed.

%%%%%%%%%%%%%%%%%%%%%%%%%%%%%%%%%%%%%%%%%%%%%%%%%%%%%%%
% Implementation Details
%%%%%%%%%%%%%%%%%%%%%%%%%%%%%%%%%%%%%%%%%%%%%%%%%%%%%%%
\subsection{Implementation Details}
\label{subsec:implementation-details}

\noindent \textbf{Network architectures:}
The ground encoder uses a frozen DINOv3-ConvNeXt-Tiny~\cite{simeoni2025dinov3} backbone with a UPerNet~\cite{xiao2018unified} head to process a $512\times512$ onboard image.
The UPerNet aggregates the $\tfrac{1}{4}$, $\tfrac{1}{8}$, $\tfrac{1}{16}$, and $\tfrac{1}{32}$ backbone feature maps.
The resulting feature map is then processed by the BEV mapper and BEV encoder to produce the final 32-dimensional BEV feature map $\mathbf{G}$, distinctiveness map $\mathbf{C}$, and frame-level uncertainty $\sigma_t$.
The BEV grid size is $224\times224$.

The aerial encoder uses the same DINOv3-ConvNeXt-Tiny backbone with a lightweight multi-scale decoder to process a $768 \times 768$ aerial image.
The decoder takes the $\tfrac{1}{4}$, $\tfrac{1}{8}$, and $\tfrac{1}{16}$ backbone feature maps, projects each scale with a $1 \times 1$ convolution, upsamples the coarser scales to the $\tfrac{1}{4}$ resolution, concatenates them, and applies a shallow convolutional head to produce the final 32-dimensional aerial feature map $\mathbf{F}$.

\noindent \textbf{Training details:}
We train the network for $25k$ iterations on $4\times$ NVIDIA Quadro RTX 6000 GPUs with a batch size of 4, using data from both the TartanDrive~2.0 and \textsc{UT-SARA-GQ} datasets.
We use the AdamW optimizer with a learning rate of $1\times10^{-4}$ and a weight decay of $1\times10^{-3}$.
We use $63$ negative poses per positive sample, sampled from translation offsets in the range [$5.0\mathrm{m}, 100.0\mathrm{m}$] and heading offsets in the range [$-90^\circ, 90^\circ$].
The InfoNCE temperature parameter $\tau$ is learnable, initialized at $0.05$ with a minimum value of $0.01$.
The distinctiveness coefficients $c_p$ and $c_m$ are set to $10.0$ and $2.0$, respectively.
The loss weights are set to $\lambda_{\text{d}}{=}1.0$ for the distinctiveness loss.

\noindent \textbf{Particle filter configuration:}
For all experiments, we use $N=128$ particles, a number chosen to balance tracking accuracy with the computational constraints of onboard deployment.
The filter is initialized around the ground-truth starting pose with Gaussian noise ($\mathrm{std}_{xy}{=}3.0\mathrm{m}, \mathrm{std}_{\theta}{=}10^\circ$).
Prediction noise is proportional to the odometry, with standard deviations set to 10\% of the measured motion.
The likelihood temperature $\tau_s$ is fixed at $1.0$.
Resampling is triggered when the effective sample size drops below 30\%.

%%%%%%%%%%%%%%%%%%%%%%%%%%%%%%%%%%%%%%%%%%%%%%%%%%%%%%%
% TartanDrive 2.0 & GQ dataset
%%%%%%%%%%%%%%%%%%%%%%%%%%%%%%%%%%%%%%%%%%%%%%%%%%%%%%%
\subsection{Q1 \& Q2: Accuracy and Generalization}
\label{subsec:q12-accuracy-generalization}

\begin{figure*}[htb]
  \centering

  \subfloat[Seen route (TD01)\label{fig:td01-traj}]{%
    \includegraphics[height=5.7cm]{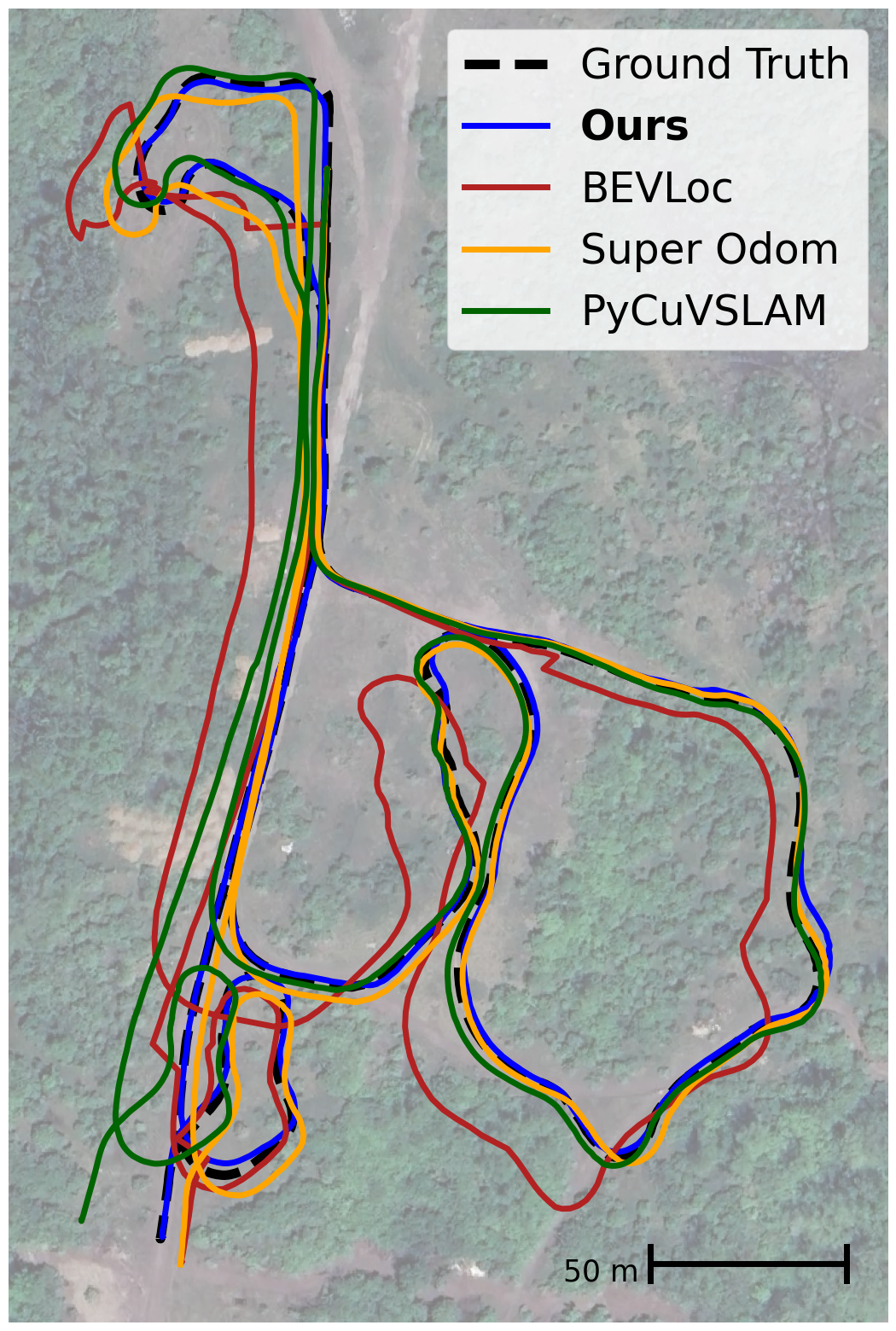}}%
  \hspace{0.3em}%
  \subfloat[Unseen route (TD10)\label{fig:td10-traj}]{%
    \includegraphics[height=5.7cm]{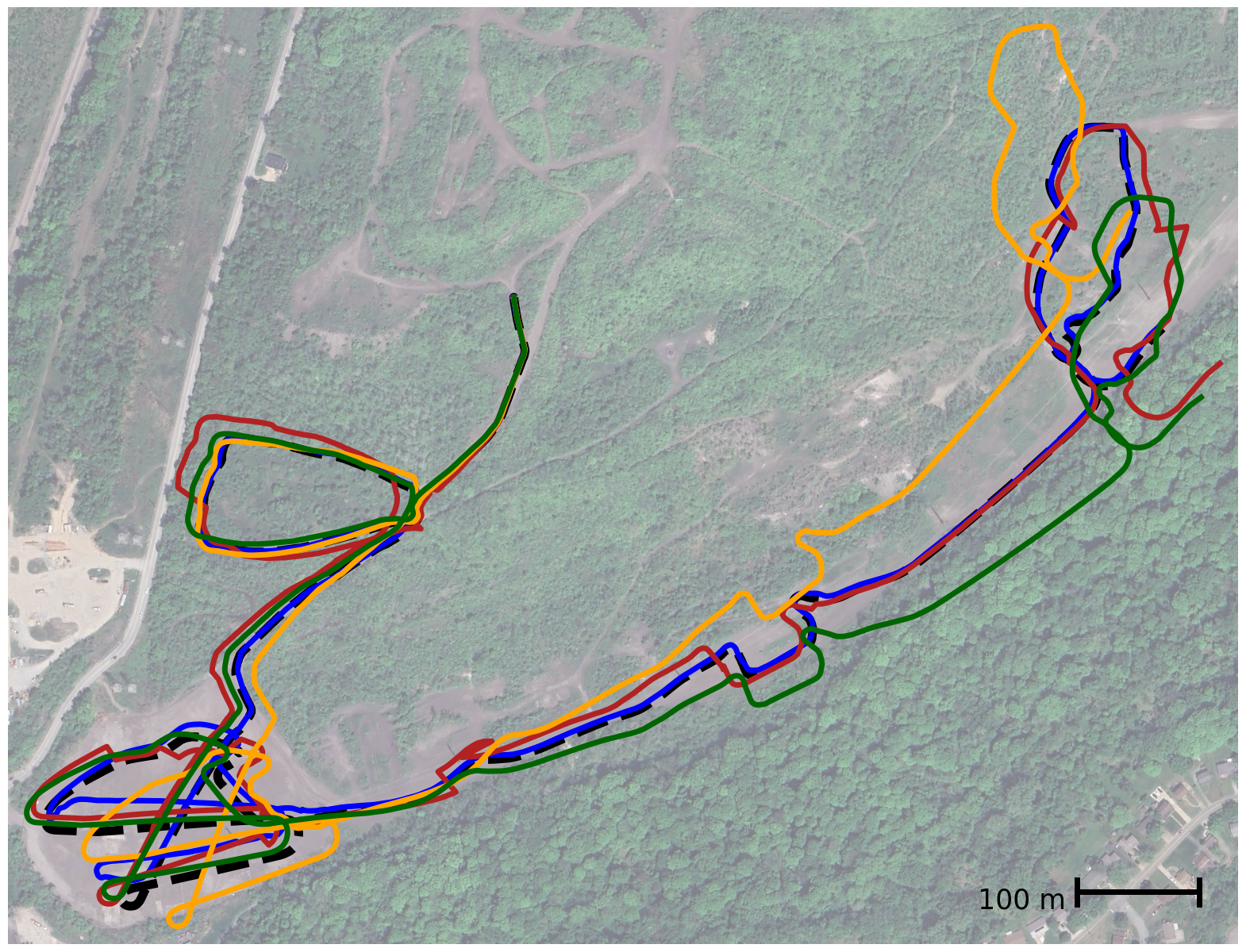}}%
  \hspace{0.3em}%
  \subfloat[Unseen route (TD22)\label{fig:td22-traj}]{%
    \includegraphics[height=5.7cm]{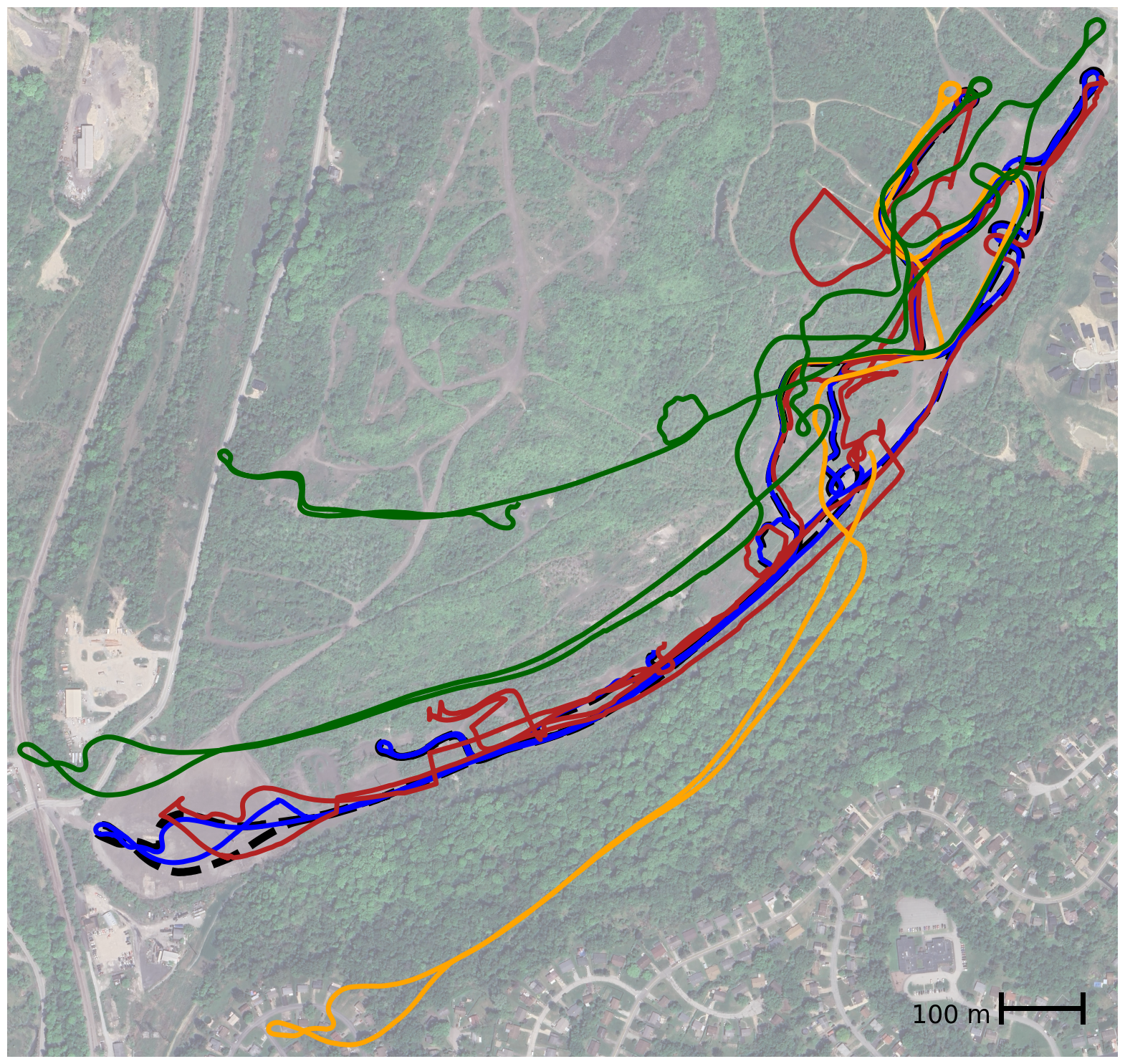}}%

  \caption{Comparison of estimated trajectories from \bpp and all baselines on the TartanDrive~2.0 dataset.}
  \label{fig:td-trajectories}
\end{figure*}

Table~\ref{tab:tartandrive_ate} presents the quantitative ATE results, while Figure~\ref{fig:td-trajectories} provides a qualitative comparison of the trajectories.
For tracking accuracy, our method achieves an average ATE of 3.10~m, significantly outperforming BEVLoc (21.90~m) and the odometry baselines.
Compared with BEVLoc, which often exhibits discontinuities when per-frame ambiguities are not fully resolved by its pose graph, our sequential filtering approach produces smooth and accurate trajectories without using absolute position fixes during inference.

\begin{figure}[htb]
    \centering
    \includegraphics[width=0.95\linewidth]{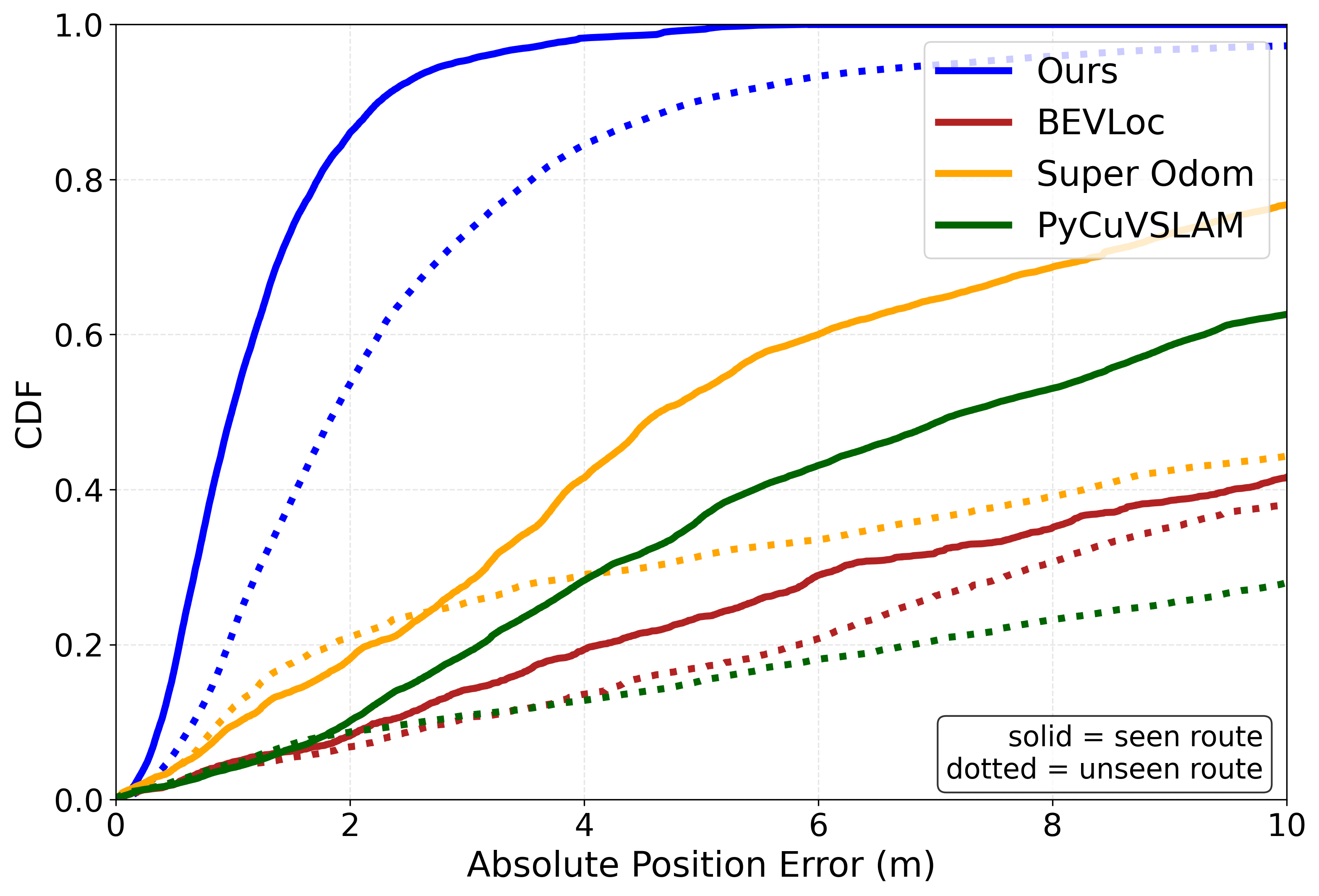}
    \caption{Cumulative distribution of absolute pose error (meters) for seen and unseen routes on the TartanDrive~2.0 dataset.}
    \label{fig:td-ape-cdf}
\end{figure}

For generalization, on routes not seen during training, \bpp maintains a low ATE of 3.61~m, again surpassing all baselines.
This demonstrates that \bpp generalizes well to novel paths while maintaining high accuracy.
The cumulative error distribution in Fig.~\ref{fig:td-ape-cdf} further confirms that our method maintains a substantially tighter error profile on both seen and unseen routes.

\input{tables/tartandrive_ate}

%%%%%%%%%%%%%%%%%%%%%%%%%%%%%%%%%%%%%%%%%%%%%%%%%%%%%%%
% UT-SARA-GQ Dataset
%%%%%%%%%%%%%%%%%%%%%%%%%%%%%%%%%%%%%%%%%%%%%%%%%%%%%%%
\subsection{Q3: Canopy and Shadow Robustness}
\label{subsec:q3-canopy-shadow}

\begin{figure}[htb]
    \centering
    \includegraphics[width=0.95\linewidth]{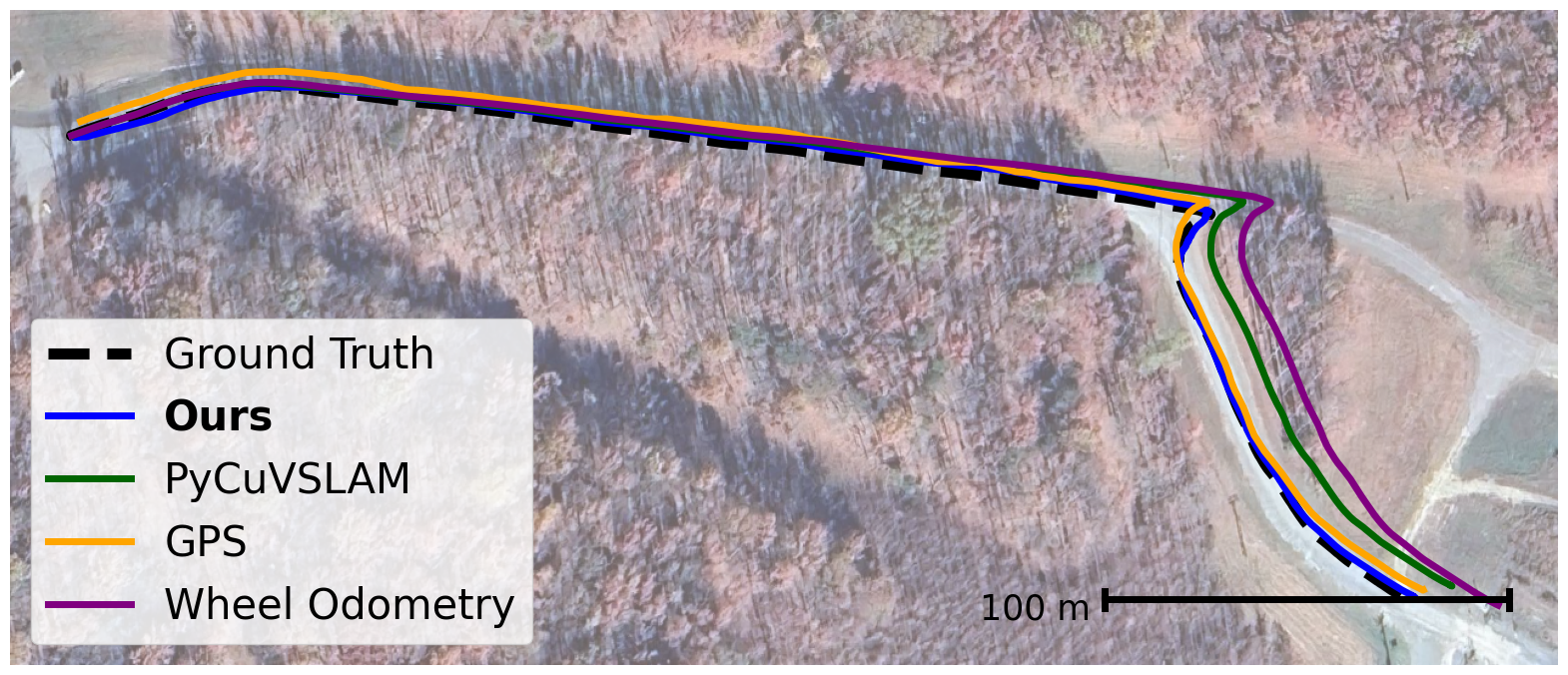}%
    
    \caption{Example trajectory (GQ01) in the \textsc{UT-SARA-GQ} dataset.}
    \label{fig:gq-traj}
\end{figure}

To test robustness under challenging aerial conditions, we use the \textsc{UT-SARA-GQ} dataset, which contains tree-canopy cover and shadowed trail segments.
As shown in Table~\ref{tab:gq-ate} and qualitatively in Figure~\ref{fig:gq-traj}, \bpp maintains track lock and estimates accurate trajectories, demonstrating that the learned features remain effective under these real-world appearance variations.

%%%%%%%%%%%%%%%%%%%%%%%%%%%%%%%%%%%%%%%%%%%%%%%%%%%%%%%
% Real-time Robot Deployment
%%%%%%%%%%%%%%%%%%%%%%%%%%%%%%%%%%%%%%%%%%%%%%%%%%%%%%%
\subsection{Q4: Real-time Performance}
\label{subsec:q4-realtime}

\begin{figure}[htb]
    \centering
    \includegraphics[width=0.95\linewidth]{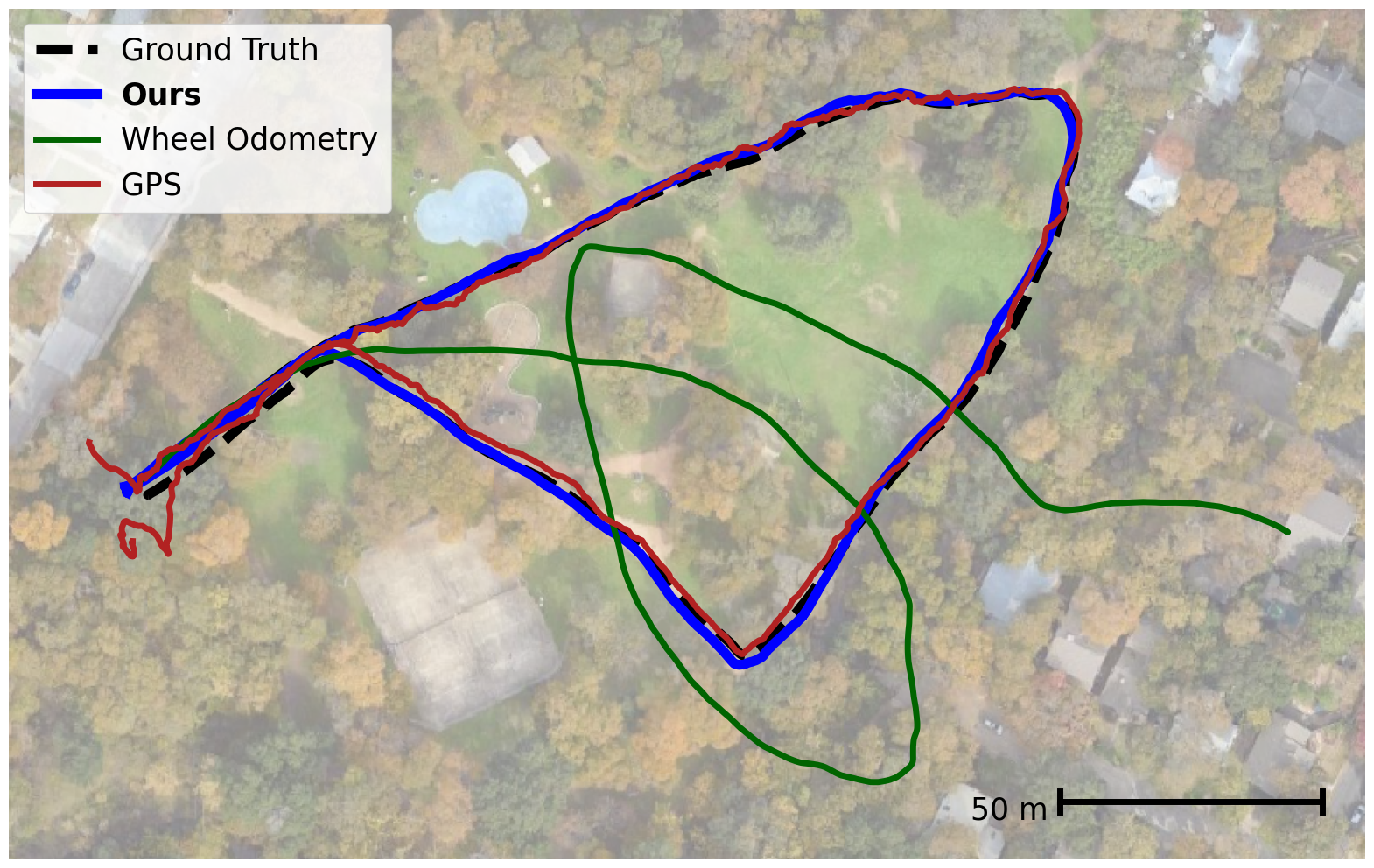}
    \caption{Trajectory from the real-time experiment (UP02) in the Urban Park. This run was manually initialized in the map GUI.}
    \label{fig:ew02-traj}
\end{figure}

We deployed our system on a Clearpath Jackal robot to evaluate its real-time performance.
The network was compiled with TensorRT~\cite{nvidia:tensorrt} and wrapped in ROS~2, achieving 10~Hz on an NVIDIA Tesla T4 GPU (see Table~\ref{tab:tensorrt-profile} for a module-level latency breakdown).
During live tests in an Urban Park, the system produced accurate trajectories using only wheel odometry for the motion-prediction step (Table~\ref{tab:gq-ate}, Fig.~\ref{fig:ew02-traj}).
For this deployment, the particle filter was initialized by manually selecting an initial pose in a GUI, demonstrating a fully GPS-free operational workflow.

\input{tables/gq_ate}

\input{tables/tensorrt_profile}

%% file: tables/tartandrive_ate.tex
\begin{table*}[htb]
  \centering
  \caption{Absolute Trajectory Error (ATE RMSE, meters) on the TartanDrive 2.0 dataset, evaluated in the UTM frame. All methods are initialized with the ground-truth pose. The table reports performance on routes seen during training (TD01–06) and unseen routes (TD07–22). Best results are shown in bold.}
  \label{tab:tartandrive_ate}
  \sisetup{detect-weight=true, mode=text, table-number-alignment=right, table-format=4.2}
  
  \resizebox{\linewidth}{!}{%
    \begin{tabular}{l *{11}{S[table-format=4.2]}}
      \toprule
      & \multicolumn{6}{c}{\textbf{Seen route (6 scenes)}} & \multicolumn{5}{c}{\textbf{Unseen route (16 scenes)}} \\
      \cmidrule(lr){2-7} \cmidrule(lr){8-12}
      \textbf{Method}
        & \multicolumn{1}{r}{TD01} & \multicolumn{1}{r}{TD02} & \multicolumn{1}{r}{TD03} & \multicolumn{1}{r}{TD04} & \multicolumn{1}{r}{TD05} & \multicolumn{1}{r}{TD06}
        & \multicolumn{1}{r}{TD07} & \multicolumn{1}{r}{TD08} & \multicolumn{1}{r}{TD09} & \multicolumn{1}{r}{TD10} & \multicolumn{1}{r}{TD11} \\
      \midrule
   
      BEVLoc~\cite{klammer2024bevloc}
        & 16.15 & 24.78 & 17.07 & 33.84 & 5.97 & 3.06
        & 23.75 & 16.63 & 17.22 & 22.69 & 26.30 \\

      PyCuVSLAM~\cite{korovko2025cuvslamcudaacceleratedvisual}
        & 8.04 & 4.61 & 16.85 & 12.08 & 2.49 & 8.26
        & 38.69 & 32.87 & 29.32 & 35.15 & 15.08 \\
        
      Super Odometry~\cite{zhao2021super}
        & 5.67 & 15.50 & 12.63 & 3.31 & 4.76 & 17.25
        & 8.76 & 16.12 & 5.11 & 54.12 & 86.38 \\

      \textbf{Ours}
        & \bfseries\tablenum{2.10}
        & \bfseries\tablenum{1.62}
        & \bfseries\tablenum{1.11}
        & \bfseries\tablenum{1.64}
        & \bfseries\tablenum{1.57}
        & \bfseries\tablenum{2.38}
        & \bfseries\tablenum{1.61}
        & \bfseries\tablenum{3.00}
        & \bfseries\tablenum{4.02}
        & \bfseries\tablenum{7.64}
        & \bfseries\tablenum{2.24} \\
    
      \bottomrule
    \end{tabular}}

  \vspace{4pt}

  \resizebox{\linewidth}{!}{%
    \begin{tabular}{l *{11}{S[table-format=4.2]}}
      \toprule
      & \multicolumn{11}{c}{\textbf{Unseen route (continued)}} \\
      \cmidrule(lr){2-12}
      \textbf{Method}
        & \multicolumn{1}{r}{TD12} & \multicolumn{1}{r}{TD13} & \multicolumn{1}{r}{TD14} & \multicolumn{1}{r}{TD15} & \multicolumn{1}{r}{TD16} & \multicolumn{1}{r}{TD17}
        & \multicolumn{1}{r}{TD18} & \multicolumn{1}{r}{TD19} & \multicolumn{1}{r}{TD20} & \multicolumn{1}{r}{TD21} & \multicolumn{1}{r}{TD22} \\
      \midrule
      
      BEVLoc~\cite{klammer2024bevloc}
        & 18.16 & 17.72 & 12.08 & 33.16 & 21.44 & 4.20
        & 27.05 & 23.88 & 25.53 & 35.38 & 55.64 \\

      PyCuVSLAM~\cite{korovko2025cuvslamcudaacceleratedvisual}   & 270.90 & 16.49 & 10.33 & 6.61 & 16.74 & 15.33
        & 42.40 & 282.43 & 57.88 & 32.86 & 163.02 \\
        
      Super Odometry~\cite{zhao2021super}
        & 344.16 & 34.91 & 285.09 & 7.63 & 4.26 & 3.82
        & 727.19 & 156.77 & 24.58 & 14.02 & 150.55 \\

      \textbf{Ours}
        & \bfseries\tablenum{5.89}
        & \bfseries\tablenum{2.72}
        & \bfseries\tablenum{3.30}
        & \bfseries\tablenum{1.93}
        & \bfseries\tablenum{2.33}
        & \bfseries\tablenum{3.17}
        & \bfseries\tablenum{3.85}
        & \bfseries\tablenum{4.69}
        & \bfseries\tablenum{2.74}
        & \bfseries\tablenum{4.23}
        & \bfseries\tablenum{4.37} \\

      \bottomrule
    \end{tabular}}
\end{table*}

%% file: tables/gq_ate.tex
\begin{table}[htb]
  \centering
  \caption{Absolute Trajectory Error (ATE RMSE, meters) on the \textsc{UT-SARA-GQ} dataset and real-time robot deployment at the Urban Park.}
  \label{tab:gq-ate}
  \sisetup{detect-weight=true, mode=text, table-number-alignment=right, table-format=4.2}
  
  \resizebox{\linewidth}{!}{%
    \begin{tabular}{l *{6}{S}}
      \toprule
      & \multicolumn{2}{c}{\textbf{Seen route}}
      & \multicolumn{2}{c}{\textbf{Unseen route}}
      & \multicolumn{2}{c}{\textbf{Real-time}} \\
      \cmidrule(lr){2-3} \cmidrule(lr){4-5} \cmidrule(lr){6-7}
      \textbf{Method}
        & \multicolumn{1}{r}{GQ01}
        & \multicolumn{1}{r}{GQ02}
        & \multicolumn{1}{r}{GQ03}
        & \multicolumn{1}{r}{GQ04}
        & \multicolumn{1}{r}{UP01}
        & \multicolumn{1}{r}{UP02} \\
      \midrule

      GPS
        & \bfseries\tablenum{3.61} & 8.19 
        & 4.61 & 5.07 
        & 2.34 & 3.45 \\

      PyCuVSLAM
        & 6.55 & 13.32 
        & 2.79 & 4.31
        & {-} & {-} \\

      Wheel Odom
        & 11.82 & 31.04 
        & 3.38 & 7.39 
        & 74.89 & 97.09 \\

      \textbf{Ours}
        & 3.68
        & \bfseries\tablenum{7.09}
        & \bfseries\tablenum{2.05}
        & \bfseries\tablenum{4.10}
        & \bfseries\tablenum{2.03}
        & \bfseries\tablenum{2.62} \\
    
      \bottomrule
    \end{tabular}}
\end{table}

%% file: tables/tensorrt_profile.tex
\begin{table}[htb]
  \centering
  \caption{Per-module inference latency (\si{\milli\second}) using TensorRT (FP16).}
  \label{tab:tensorrt-profile}
  
\resizebox{\linewidth}{!}{%
  \begin{tabular}{@{}l *{7}{S[table-format=2.2]}@{}}
    \toprule
    & \multicolumn{1}{c}{\makecell{Ground\\Encoder}}
    & \multicolumn{1}{c}{\makecell{BEV\\Mapper}}
    & \multicolumn{1}{c}{\makecell{BEV\\Encoder}}
    & \multicolumn{1}{c}{\makecell{Aerial\\Encoder}}
    & \multicolumn{1}{c}{\makecell{Patch\\Sampler}}
    & \multicolumn{1}{c}{\makecell{Scoring\\Head}}
    & \multicolumn{1}{c}{\makecell{Total\\(+I/O)}}
    \\
    \midrule
    \textbf{RTX 3080} & 3.85 & 3.82 & 1.17 & 3.96 & 2.91 & 5.80 & 34.96 \\
    \textbf{Tesla T4} & 10.75 & 11.64 & 3.47 & 11.04 & 9.84 & 27.66 & 92.36 \\
    \bottomrule
  \end{tabular}
}

\end{table}

%% file: content/conclusion.tex
\section{Conclusion, Limitations, and Future Work}
\label{sec:conclusion}

This paper presents \bpp, a sequential cross-view geo-localization system that integrates a particle filter with a learned observation model.
By scoring continuous pose hypotheses through BEV--aerial feature matching, \bpp provides accurate vision-only global localization in the UTM frame.
Across real-world off-road experiments, the method consistently outperformed odometry and retrieval-based baselines, while remaining sufficiently efficient for real-time onboard deployment.

A current limitation is that \bpp performs best within the training distribution, where route semantics, trail geometry, and aerial appearance are similar to those represented in the training data.
Performance tends to degrade in out-of-distribution environments, especially when routes are semantically different from the training set, when aerial and ground appearances differ substantially, or when the traversable route is too narrow to be reliably resolved in the aerial imagery.
In these cases, the observation model becomes less informative, leading to higher uncertainty and lower localization accuracy.

Future work will focus on improving robustness to out-of-distribution conditions by training on larger and more diverse datasets with wider variation in cameras, capture heights, and environments, such as the Mapillary Street-level Sequences Dataset~\cite{warburg2020mapillary}.
We also plan to localize against live drone imagery to mitigate aerial-map mismatch and to incorporate observability-aware path planning that favors routes expected to remain confidently localizable by \bpp.